\documentclass[10pt,twocolumn,letterpaper]{article}

\usepackage{cvpr}              % To produce the CAMERA-READY version
\definecolor{cvprblue}{rgb}{0.21,0.49,0.74}
\usepackage[pagebackref,breaklinks,colorlinks,allcolors=cvprblue]{hyperref}
\usepackage{graphicx}
\usepackage{multirow}
\usepackage{makecell}
\usepackage{colortbl}

\def\paperID{11011} % *** Enter the Paper ID here
\def\confName{CVPR}
\def\confYear{2025}

\title{NoisyEQA: Benchmarking Embodied Question Answering Against Noisy Queries}

\author{Tao Wu$^1$ \quad Chuhao Zhou$^1$ \quad Yen Heng Wong$^1$ \quad Lin Gu$^2$ \quad Jianfei Yang$^1$\\
$^1$ MARS Lab, Nanyang Technological University \quad $^2$ RIKEN AIP \\
{\small \tt \{twu019@e., chuhao002@e., wong1348@e., jianfei.yang@\}ntu.edu.sg lin.gu@riken.jp}  }

\begin{document}
\twocolumn[{
\renewcommand\twocolumn[1][]{#1}
\maketitle
\begin{center}
    \captionsetup{type=figure}
    \includegraphics[width=0.88\textwidth]{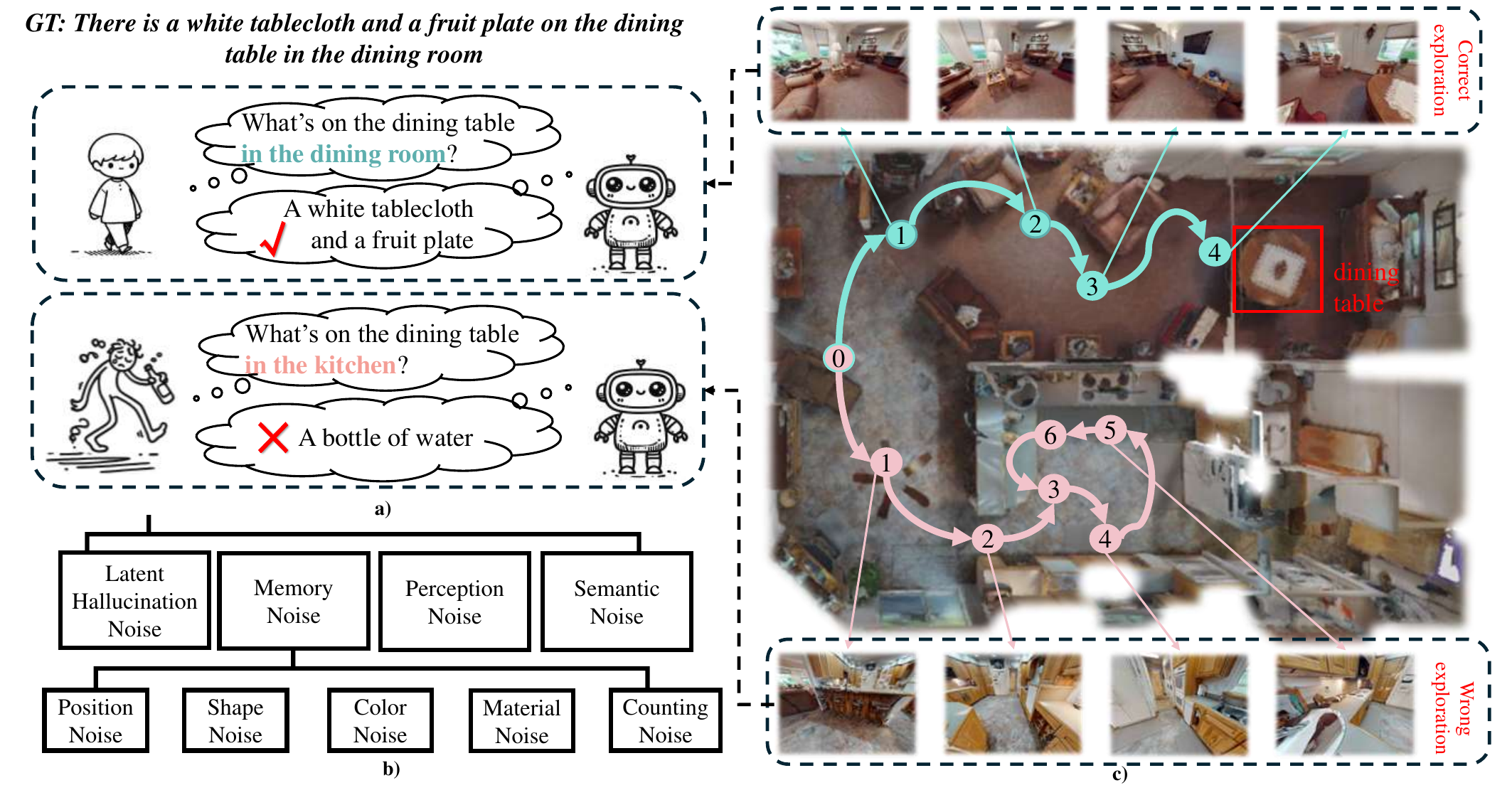}
    \captionof{figure}{(a) A dining table locates in the dining room, a drunk person mistakenly recalls its location, causing the Agent incorrectly makes a fabricated answer. (b) A systematic taxonomy for human noisy questions. (c) Comparison between Agent’s exploration paths under the accurate question(Green) and noisy question(Pink), the latter may be trapped in a loop, causing physical damage.}
    \label{Figure 1}
\end{center}
}]
\maketitle

\begin{abstract}
The rapid advancement of Vision-Language Models (VLMs) has significantly advanced the development of Embodied Question Answering (EQA), enhancing agents' abilities in language understanding and reasoning within complex and realistic scenarios. However, EQA in real-world scenarios remains challenging, as human-posed questions often contain noise that can interfere with an agent's exploration and response, bringing challenges especially for language beginners and non-expert users. To address this, we introduce a \textbf{NoisyEQA} benchmark designed to evaluate an agent’s ability to recognize and correct noisy questions. This benchmark introduces four common types of noise found in real-world applications—Latent Hallucination Noise, Memory Noise, Perception Noise, and Semantic Noise—generated through an automated dataset creation framework. Additionally, we also propose a 'Self-Correction' prompting mechanism and a new evaluation metric to enhance and measure both noise detection capability and answer quality. Our comprehensive evaluation reveals that current EQA agents often struggle to detect noise in questions, leading to responses that frequently contain erroneous information. Through our Self-Correct Prompting mechanism, we can effectively improve the accuracy of agent answers.
\end{abstract}    
\section{Introduction}
\label{sec:intro}
Embodied Question Answering (EQA) is a task where an agent actively explores a scene, builds a comprehensive understanding of the environment, and answers questions about it using natural language~\cite{das2018embodied,liu2024aligning,ma2022sqa3d,majumdar2024openeqa}. Unlike Visual Question Answering (VQA), which relies on a single static image~\cite{zhang2023learning,antol2015vqa,wang20243d,li2024configure,shao2023prompting}, EQA requires responses based on multiple images collected during the agent's exploration.

Existing QA systems often assume that the questions posed to them are clear and accurate, leading to challenges when erroneous human presumptions or biases are present. This issue is concerning for EQA, as it must interact with diverse clients and conditions in the real world, where incorrect responses can frustrate or even harm users. The situation is more severe in critical or high-stakes environments.
%For example, in critical or high-stakes environments like a manufacturing workshop, incorrect responses potentially cause physical harm or safety risks~\cite{norman1983design}. 
For instance, when assisting an Alzheimer’s patient, the agent  might be asked to locate a commonly \textbf{misplaced object}, such as glasses. If the agent repeatedly searches this area based on the patient’s erroneous presumptions, it not only becomes inefficient but may also collide with obstacles or even the patient, posing serious safety risks. Unlike ideal scenarios with accurate questions, real-world queries often contain ambiguities and inaccuracies. It highlights the need for EQA systems to flexibly handle these noisy questions, thereby enhancing practicality for real-world applications.

\begin{figure}[h!] 
    \centering % 图片居中
    \includegraphics[width=0.45\textwidth]{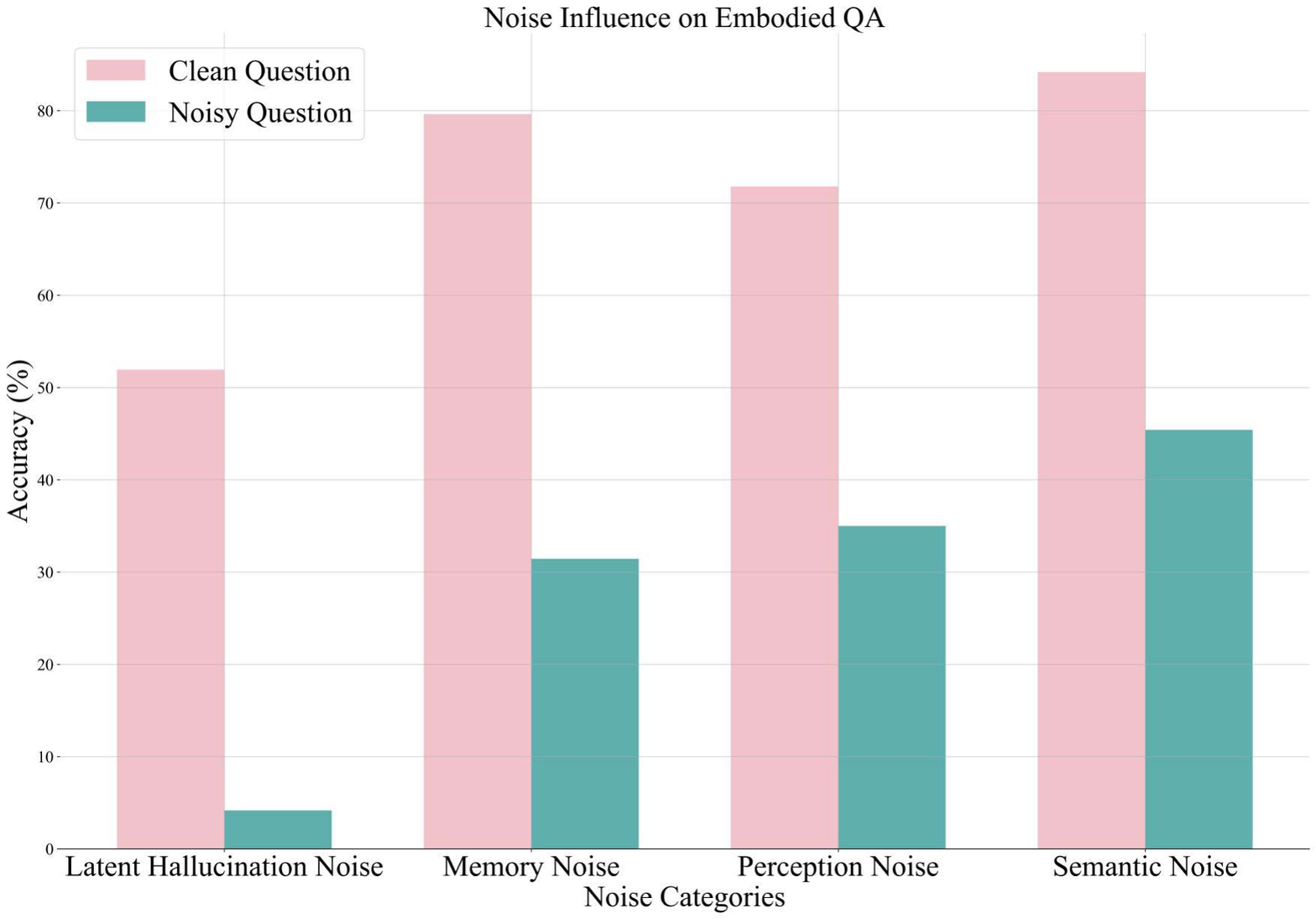} 
    \caption{The accuracy of Explore-EQA decreases significantly on noisy questions compared to clean questions.} % 图片标题
    \label{fig:Figure 8}
\end{figure}

To understand the limitations of existing EQA systems in handling real-world ambiguities, we examined \textbf{Explore-EQA}~\cite{ren2024explore} and \textbf{OpenEQA}~\cite{majumdar2024openeqa}, which demonstrate weaknesses due to noisy queries. As shown in Figure \ref{Figure 1} (a) and the pink path in (c), when faced with a noisy question from a drunk person, 
%the agent not only needs to spend more resources and effort exploring but may also repeatedly search the same location and produce incorrect responses. 
the agent may repeatedly search the same location and produce incorrect responses, results in useless exploration or even physical damage. The quantitive results in Figure \ref{fig:Figure 8} further demonstrates our advocation. Here, we treat the “noise” as inaccuracies and ambiguities within user queries, which hinder the agent to accurately interpret and respond. The results highlight a critical gap in current EQA systems' capabilities to handle clean and noisy questions. We find that these noisy questions can be derived from the concept of “Presumptuous Questions” where queries containing incorrect or unfounded assumptions about the environment. It is the misleading premises that introduced by such presumptions that induce agents to make erroneous responses. Therefore, we advocate that it is significant to systematically model and define these noisy questions, which facilitates the advent of noise-robust agents.

%thereby illustrating the need for EQA systems capable of critically assessing and managing assumptions inherent in real-world interactions.

\iffalse In addition to methodological limitations, existing evaluation metrics also fall short in assessing EQA systems' performance on noisy or assumptive questions. The multiple-choice format employed by Explore-EQA~\cite{ren2024explore} is not well-suited for real-world applications, as agents do not encounter pre-set answers when responding to human queries. OpenEQA~\cite{majumdar2024openeqa} introduces LLM-Match to evaluate open-ended responses, which partially addresses this limitation. However, when faced with suggestive or leading questions from humans, LLM-Match may still assign a perfect score, even if the agent's response includes inaccurate descriptions. This scoring bias significantly hinders the effective assessment of noise detection abilities. \fi

Hence, we introduce a novel multimodal benchmark, \textbf{NoisyEQA}, specifically designed to assess agents’ performance in handling noisy questions. 
%First, we analyze and categorize common types of errors observed in human-posed questions, identifying typical noise patterns that arise from clients’ assumptions and implicit biases. 
With analysis of typical noise pattern observed in human-posed questions, we model and define them into 4 types, as shown in Figure \ref{Figure 1} (b). Specifically, latent hallucination noise arises when a question refers to non-existent objects or entities in a given scene; perception noise originates from the inaccurate object descriptions caused by visual disturbances or occlusions; semantic noise arises from errors in interpreting environment where the original objects are substituted by semantically related but incorrect ones; and memory noise is related to memory error where the object attributes recalled by humans mismatch their actual attributes. We further subdivide memory noise into five attributes—location, color, material, count, and shape—capturing common concepts of mistaken presumptions. With these definitions, we design a systematic framework powered by a LLM to automatically generate noisy questions and ultimately 500 noisy questions are incorporated in the NoisyEQA benchmark.

Furthermore, we take the first step in enhancing the robustness of EQA agents towards noisy questions. A simple yet effective \textbf{Self-Correction} mechanism is proposed, which can be treated as plug-ins and seamlessly incorporated into any EQA system. The core motivation of our Self-Correction mechanism is to prompt agents to inspect the noise by comparing their visual observations with human questions. Once noises is detected, the agents is asked to provide a clear explanation of the noise source, based on which the noise-free responses can be generated. Notably, the self-correction mechanism can be easily extended into a human-in-the-loop approach, where the corrected responses from agents further confirmed by humans before execution. In this case, our proposed method sufficiently meet the AI Act’s transparency requirements~\cite{edwards2021eu,verde2024seizing}, which is critical for embodied systems.

Finally, we argue that existing metrics (e.g., accuracy) is incapable of evaluating agents' performances to noisy questions. Therefore, a novel evaluation framework that specially designed for noisy questions is proposed. To be specific, we elaborately design a evaluation scale ranging from 1 to 5 to assess agents in terms of noise detection, noise correction and accurate generation. To fully demonstrate the fine-grained capabilities of agents in handling noises, two noise-related metrics—Detection Rate (DR) and Correction Rate (CR)—are also introduced to specify whether the agents achieve to detect and correct noise lied in the questions. The combination of these metrics provide a framework that can comprehensively evaluate the agents' performance in real-world EQA tasks.

In summary, our contributions lied in 4 folds:
\begin{enumerate}
\item We propose the \textbf{NoisyEQA} benchmark, a benchmark dataset with diverse noise types to evaluate the robustness of EQA agents against noisy questions
\item We develop a novel \textbf{LLM-powered framework} to systematically generate noisy questions across various types, enabling controlled and scalable data generation
\item We introduce a \textbf{Self-Correction} mechanism, a method meets AI transparency, to detect and correct the noisy questions, allowing to generate noise-free responses.
\item We propose an innovative \textbf{evaluation framework} that specially assesses agents' ability to handle noisy questions, allowing to comprehensively evaluate the agents’ performance in real-world EQA tasks.
\end{enumerate}

\section{Related Work}
\subsection{Embodied Question Answering}

The Embodied Question Answering (EQA) task aims to navigate a robot agent to an appropriate location in real-world scenarios and enable question-answering through physical interaction with humans. Existing methods for EQA can be divided into two categories: Traditional Neural Networks and those utilizing Large Language Models (LLMs)/Vision-Language Models (VLMs)~\cite{liu2024aligning}.

\textbf{Traditional Neural Network.} 
Das\emph{et al.}~\cite{das2018embodied} first defined the EQA as a task requiring agents to navigate and answer questions in a real environment. To address this task, most early works employ traditional neural networks with relatively small scales. For example, Das \emph{et al.}~\cite{das2018neural} proposed a modular, hierarchical strategy for learning long-range navigation from language input. Yu \emph{et al.}~\cite{yu2019multi} further extended EQA by addressing questions involving multiple targets.

\textbf{LLMs and VLMs.} 
Leveraging the strong reasoning and generalization capabilities of large-scale LLMs and VLMs, EQA has witnessed significant advancements, particularly in complex reasoning, memory integration, and adaptive exploration capabilities~\cite{dorbala2024s,majumdar2024openeqa,ren2024explore}. Based on LLMs and VLMs, more complex EQA tasks that are similar to real application can be achieved. For instance, OPEN-EQA~\cite{majumdar2024openeqa} focus on EQA tasks with episodic memory (EM-EQA) and active exploration (A-EQA). Based on Open-EQA, agents can memorize history observations and exploration by LLMs equipped with scene graphs, enabling adaptively recall them to answer queries. To explore the real environment more efficiently, Ren \emph{et al.}~\cite{ren2024explore} leveraged an external semantic map to highlight regions worth investigating, optimizing agents' exploration process by minimizing unnecessary actions.

\subsection{Learning with Noisy Labels}
Facilitated by availability of large datasets, deep learning has achieved remarkable success across numerous domains. However, the scarcity of high-quality data remains a significant concern in many real-world applications. To develop robust models that can effectively learn from noisy training data, many methods has been developed for robust learning with noisy labels~\cite{northcutt2021confident,goh2023activelab,zhang2024badlabel,yuan2024early,sheng2024adaptive,wan2024unlocking,quinonero2022dataset}. For example, Quinonero \emph{et al.}~\cite{quinonero2022dataset} alleviated the distribution shifts between training and testing data, enhancing model adaptability in real-world scenarios. In the field of EQA, Luo \emph{et al.}~\cite{luo2023robust} first proposed a noise-robust model that employs a hierarchical approach in the navigation module to filter unreliable labels, along with dual-branch training in the VQA module to enhance the performance in noisy environments. Instead of only considering the noise in training data, Northcutt \emph{et al.}~\cite{northcutt2021pervasive} highlighted that noise in testing data is equally critical and should not be overlooked. Inspire by this, our approach takes the first step in addressing the noise in testing data specifically for EQA tasks, and focuses on alleviating noise that will be encountered in real-world EQA tasks.

\begin{figure*}[ht!] 
    \centering % 图片居中
    \includegraphics[width=1\textwidth]{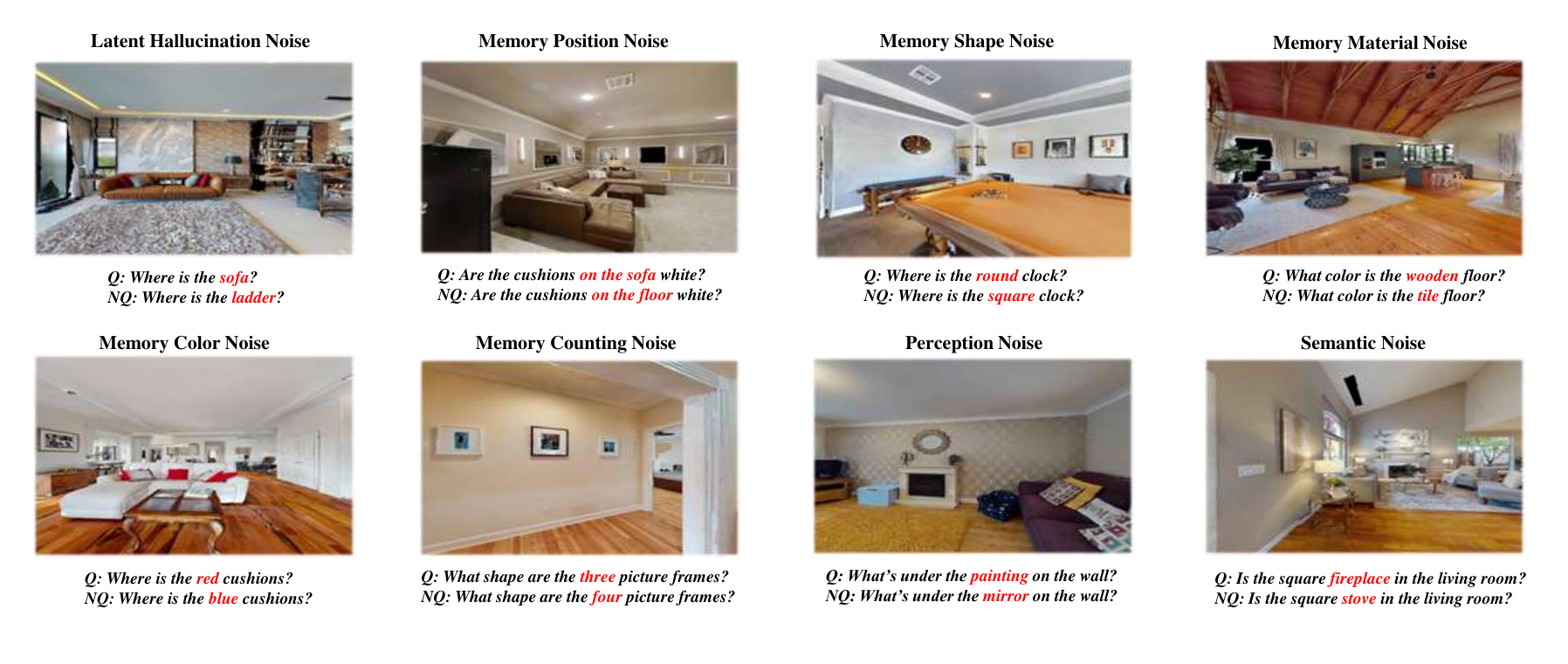} 
    \caption{Noise types in NoisyEQA benchmark: \textbf{Latent Hallucination Noise}, \textbf{Memory Noise}, \textbf{Perception Noise}, and \textbf{Semantic Noise}. Memory Noise is further subdivided into \textbf{Memory Position Noise}, \textbf{Memory Shape Noise}, \textbf{Memory Color Noise}, \textbf{Memory Counting Noise}, and \textbf{Memory Material Noise}.} % 图片标题
    \label{fig:noise-type}
\end{figure*}
\subsection{Information Inconsistency}
\textbf{Information inconsistency} is a tricky problem faced by existing Vision-Language Models (VLM). Hallucinations, where the generated language output does not align with the visual input~\cite{guan2024hallusionbench,wang2024mitigating,jiang2024hal,liu2024survey}, is one of the typical forms of \textbf{information inconsistency}.
Oceans of works are proposed to alleviate the hallucinations in VLMs~\cite{yu2024hallucidoctor,hao2024quantifying,jiang2024hallucination,gunjal2024detecting,leng2024mitigating}. For example, Li \emph{et al.}~\cite{li2023evaluating} formed hallucination detection as a binary classification problem, aiming at eliminating the misalignment of objects between agent responses and visual cues. However, existing works predominantly concentrate on hallucinations generated by agents and overlook one of the most common types of information inconsistency: misalignment between user instructions and the visual cues. Some researchers have tried to formulate this issue. For instance, Taioli \emph{et al.}~\cite{taioli2024mind} introduced a novel benchmark for Vision-Language Navigation in Continuous Environments (VLN-CE) that simulates realistic instruction errors. Followed by a cross-modal transformer to detect these errors, robustness against inaccurate human instructions is enhanced. Gao et al.~\cite{gao2025dissecting} also introduced the Self-Contradictory Instructions benchmark to evaluate agents in recognizing conflicting commands, and they proposed the Cognitive Awakening Prompting enhance the detection of such noisy instructions. These methods address specific inconsistency issues, however, a framework for systematically analyzing noise types in human inputs is absent. Our work fills this gap by defining a thorough taxonomy on noise types within human questions from a human-agent interaction perspective, comprehensively modeling and solving the information inconsistency in EQA tasks.

%-------------------------------------------------------------------------

\section{Dataset}
\subsection{Overview}

When querying about agents, humans often unconsciously introduce various presumptions and implicit biases, which can hinder the agents from generating accurate responses.
% especially for those who are struggling for mental issues or language barriers.
To model these presumptions and implicit biases in practical scenarios, we define them as different types of ``Noise" and taxonomize them into two main categories: \textbf{Latent Hallucination Noise} and \textbf{Active Noise}. The primary difference between these two types of noise lies in the presence of factual errors in human questions. \textbf{Active Noise} arises from the factual errors caused by humans' misperception or misunderstanding of the environment during interacting with agents. \textbf{Latent Hallucination Noise} occurs when questions refer to non-existing objects or entities within the scene.

In this work, we aim to model such mistaken presumptions from humans with finer granularity. As shown in Figure \ref{fig:noise-type}, we subdivide the \textbf{Active Noise} into three subtypes: \textbf{Memory Noise}, \textbf{Perception Noise}, and \textbf{Semantic Noise}. Besides, \textbf{Memory Noise} is further classified into five categories based on the forgetting of different object attributes: Position, Counting, Shape, Color and Material.
% not only memory noise, other noise also contain 5 sub categories?

\begin{figure}[ht!] 
    \centering % 图片居中
    \includegraphics[width=0.5\textwidth]{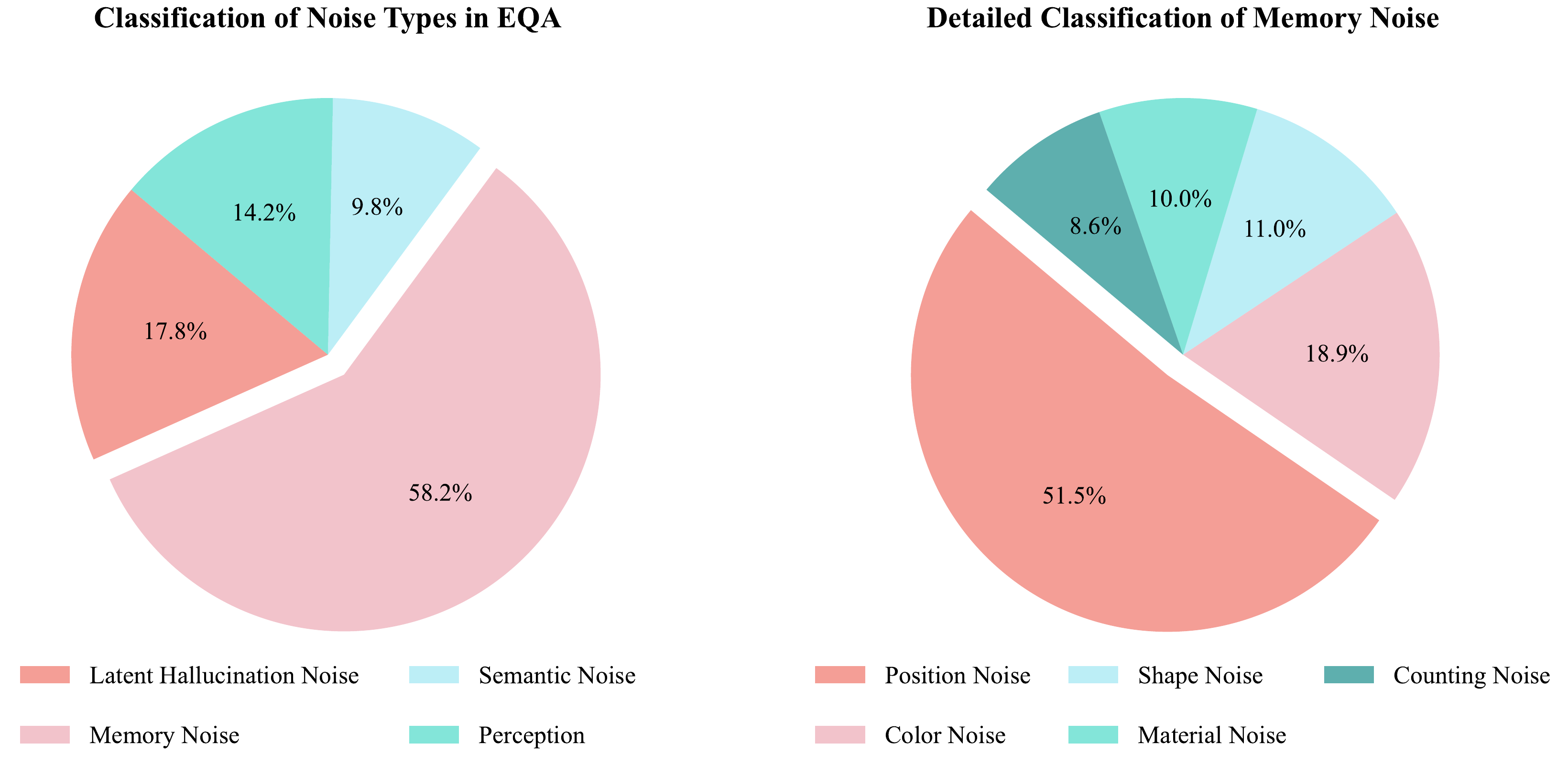} 
    %\caption{(a) shows the distribution of various types of noisy questions. (b) presents the distribution of the attributes within Memory Noise} % 图片标题
    \caption{(a) Distribution of various types of noisy questions. (b) Distribution of the attributes within Memory Noise.} % 图片标题
    % \label{fig:Figure 3}
    \label{fig:distribution_of_noisy_EQA}
\end{figure}

In total, we generate 500 noisy questions corresponding to 4 types of noise. The distribution of ``noise types" and the ``attributes of \textbf{Memory Noise}" is illustrated in Figure \ref{fig:distribution_of_noisy_EQA} (a) and (b), respectively. Please refer to supplementary materials for more details about the noise types distributions.

%\subsection{Detailed Classification}
\subsection{Definition of Noise Types}
In this subsection, we elaborate our definition of different types of noises in the NoisyEQA dataset.

\textbf{Latent Hallucination Noise} refers that questions from human contain non-existent objects in the scene. Such non-existent objects could trigger the hallucination of the agent, causing it to incorrectly generate fabricated responses that incorporate non-existent items. This kind of noise is common in the tasks of scene understanding. For example, when asked about the location of a non-existent object, the agent tends to fabricate a position rather than indicate that the object does not exist in the scene. This not only compromises EQA accuracy but also severely misleads users in practical applications.

\textbf{Memory Noise} arises from memory errors, where the human inaccurately recalls an attribute of an object, results in a mismatch between memory and reality. Imagine a scene with a white vase in the living room. If a human incorrectly recalls the vase as being in the bedroom, an example of memory noise about location attribute could be ``Is the vase in the bedroom white?". For other attributes, similar noise could be introduced into the questions.

\textbf{Perceptual Noise} originates from inaccuracies in humans' perception of the environment, simulating the visual ambiguities that they may encounter when observing objects in a scene. For instance, a nearsighted person without glasses may misidentify the object A as another incorrect object, B, in the scene. In this case, a question containing perceptual noise could be ``What's the shape of object B?".

\begin{figure}[t!] 
    \centering % 图片居中
    \includegraphics[width=0.45\textwidth]{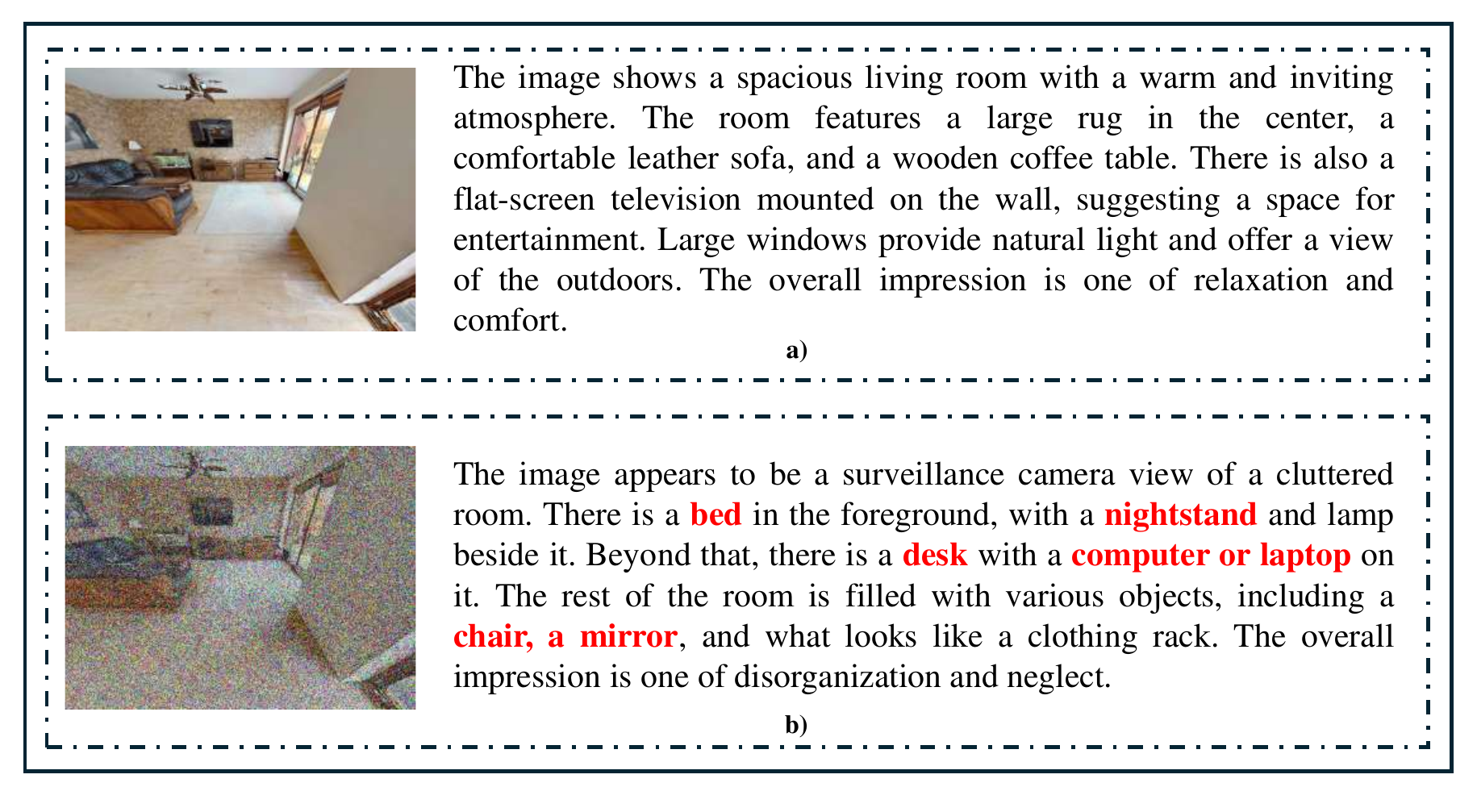} 
    \caption{(a) original image and description and (b) disturbed image and description.} % 图片标题
    \label{fig:perceptual-noise}
\end{figure}

\textbf{Semantic Noise} is caused by the incorrect substitutions of objects in a scene when humans attempt to describe the environment. Unlike perceptual noise, which involves visual misidentification of objects, semantic noise contains logically feasible substitutions with semantically related but incorrect objects. Intuitively, in a living room, substituting ``fireplace" with a semantically related object (e.g., ``stove") could introduce such noise.

\subsection{Paradigms of Question Generation}
\begin{figure}[t!] 
    \centering % 图片居中
    \includegraphics[width=0.45\textwidth]{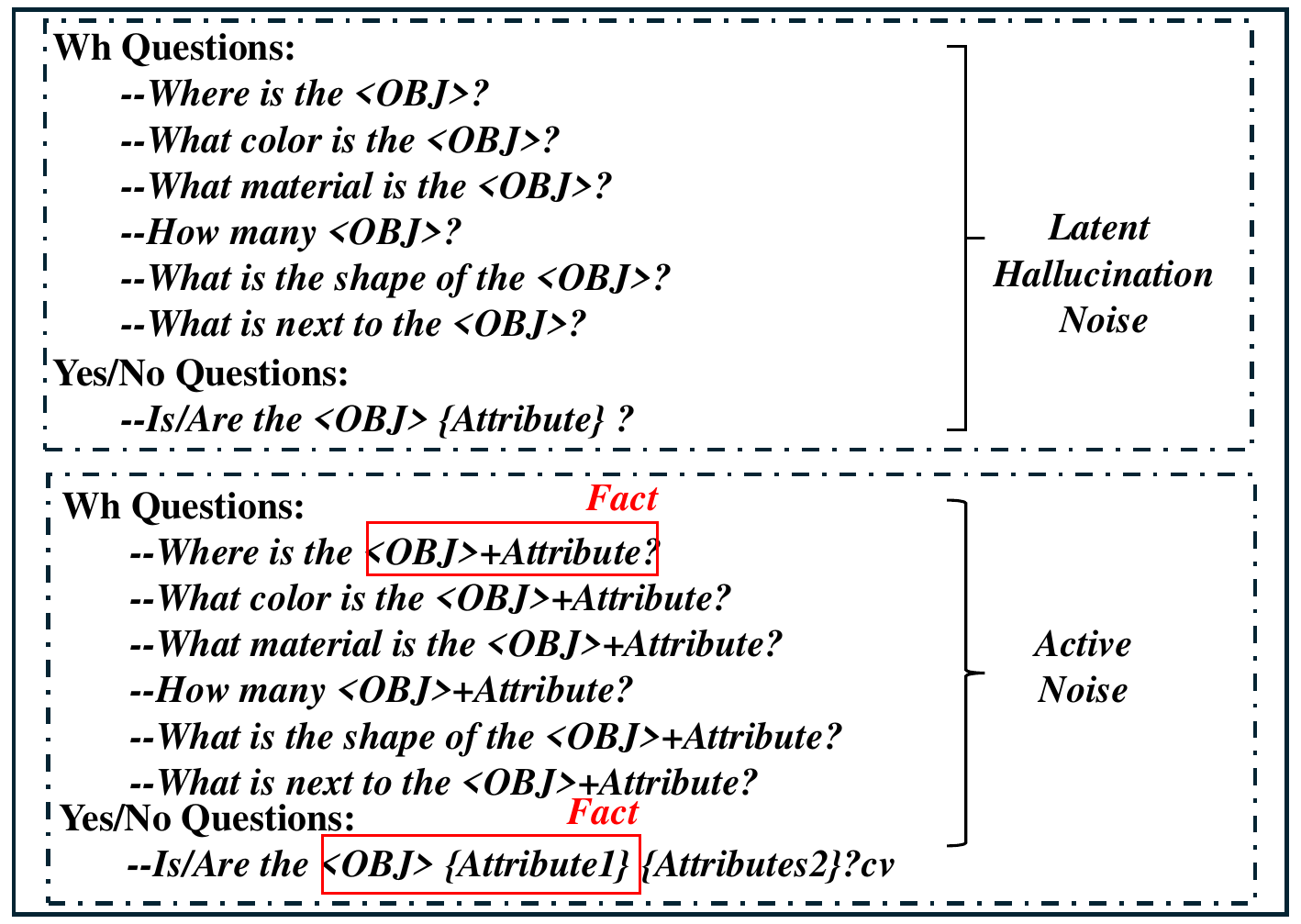} 
    \caption{Different question templates for two main noise types.} % 图片标题
    \label{fig:question-generation}
\end{figure}

Based on our definitions of different types of noise, we present a systematic framework powered by structured templates and LLMs to automatically generate noisy questions, as shown in Figure \ref{fig:question-generation}. Inspired by OpenEQA~\cite{majumdar2024openeqa}, we focus on generating open-ended question-answer pairs to better simulate the challenges of real-world environments.

For \textbf{Latent Hallucination Noise}, we use both Wh-questions and yes/no-questions templates, such as ``\texttt{Where is the <OBJ>?}" and ``\texttt{Is the <OBJ> \{Attribute\}?}". %with non-existent objects \texttt{<OBJ>} to induce hallucinated responses from the model. 
The non-existent objects \texttt{<OBJ>} and the \texttt{\{Attribute\}} aim to induce hallucinated responses and simulate the erroneous human presumptions, respectively. Given a scene, we first establish a list of all present objects, then introduce a non-existent object with a random attribute to form a noisy question. For example, if a living room contains a table, chair, and book, one of the generated questions with latent hallucination noise could be ``Where is the vase in the living room?" or ``Is the vase in the living room white?" Although this question seems to be reasonable, it introduces a phantom object (vase), increasing the likelihood of hallucinated responses.

As for \textbf{Active Noise}, the \textbf{facts} (red boxes in Figure \ref{fig:question-generation}) are pre-defined by assigning each object known attributes. We then introduce different types of Active Noise by modifying or substituting either the object or its attribute. In this way, \textbf{factual errors} between the pre-defined facts (or scenes) and the noisy questions are introduced. Except for Wh-questions, a yes-no question template is also utilized to introduce additional attributes (\textit{\{Attribute2\}}), simulating the erroneous presumptions from humans on object attributes.
By this way, noisy questions that contain both factual errors and human erroneous presumptions can be generated.

Concretely, for \textbf{Memory Noise}, we first select an object with its original attributes from a scene, then replace the attribute with an incorrect one to introduce noise. For instance, to create memory noise related to the location attribute, we select a vase that located in the `living room' and change its location to the `bedroom', then ask ``What color is the vase located in the bed room?" or ``Is the vase located in the bed room white?". This method can similarly be applied to other attributes, allowing us to comprehensively evaluate the robustness against various recall errors. 

To generate \textbf{Perceptual Noise}, we add Gaussian noise to the original image of the scene to simulate perceptual confusion. Subsequently, a Vision-Language Model (VLM)~\cite{achiam2023gpt} is utilized to generate a description of the disturbed image, as shown in Figure \ref{fig:perceptual-noise}, some objects are incorrectly described due to the distortion (marked by \textcolor{red}{red}). We then construct questions with perceptual noise based on these misidentified objects. According to Figure \ref{fig:perceptual-noise}, the sofa near the television is recognized as a bed, and a noisy question is formulated as ``What's the color of the bed near the television?" or ``Is the bed near the television made of wood?".

Finally, we generate \textbf{Semantic Noise} by substituting the object in the scene with semantic-related but incorrect object. Particularly, objects from both OpenEQA~\cite{majumdar2024openeqa} and HMEQA~\cite{ren2024explore} datasets are gathered, and BERT is adopted to embed all candidate objects. We then create a pair-wise similarity matrix to indicate the semantic correlations among objects. For an original object in each question, we first select the objects with top-10 similarities to form a subset of candidates. Then, the logically related object is further chosen from the subset by humans according to two principles: category overlap or functional association. Specifically, category overlap is defined by widely accepted taxonomy, for example, `sofa' and `armchair' are both belonged to `furniture'. Functional association considers objects with similar roles in the same scene, for example, `fireplace' and `stove' both support for heat function.

Consequently, we replace the original object with a logically related one to introduce the semantic noise. Take the scene of living room as an example, an question with semantics noise could be ``What material is the stove in the living room?" or ``Is the stove in the living room silver?" while the ``stove" is a ``fireplace" in reality.

After generating the dataset, we recruit two dedicated volunteers to conduct a thorough review and validation. The volunteers meticulously examine each entry for inconsistencies, errors, and potential biases. Additionally, they also verify that the generated data accurately contains the intended noise types across various scenarios.

\section{Self-Correction Mechanism}
Once the presumptions and implicit biases are incorporated into the human questions (noisy questions), the accuracy of responses generated by agents will significantly degrade. To address this issue, we propose a Self-Correction mechanism to enhance the robustness of agents against noisy questions. 

As illustrated in Figure \ref{fig:self-correction} (a), for a scene with multiple views, the noisy question is marked by blue. Instead of directly answering the question, we design an additional ``confidence question" in yes/no format (which is marked by green) for the agent to answer first. By taking the probability regards to ``yes" token as confidence score, the agent could identify the most reliable view from the scene. Then, the agent equipped with our Self-Correction mechanism answers the noisy question based on the selected view. The Self-Correction mechanism has two formats: \textbf{Noise Aware Prompt (NAP)} and \textbf{Noise Aware Chain of Thought (NACoT)}, which are elaborated in the following.

\subsection{Noise Aware Prompt (NAP)}
The Noise Aware Prompt (NAP) adopts an extra cognition prompt to enhance agents' sensitivity towards potential noise in human questions. As shown in Figure \ref{fig:self-correction} (b), a simple yet effective prompt is appended to the input of the agent, which reminds the agent to carefully review the question-related information before generating a response. Instead of altering the reasoning process of the agent, NAP encourages the agent to aware the mismatches between human questions and visual clues. By this way, the agent achieves to identify the noisy question more accurately, resulting in more robust and reliable responses.

\subsection{Noise Aware Chain of Thought (NACoT)}
\begin{figure}[t!] 
    \centering % 图片居中
    \includegraphics[width=0.5\textwidth]{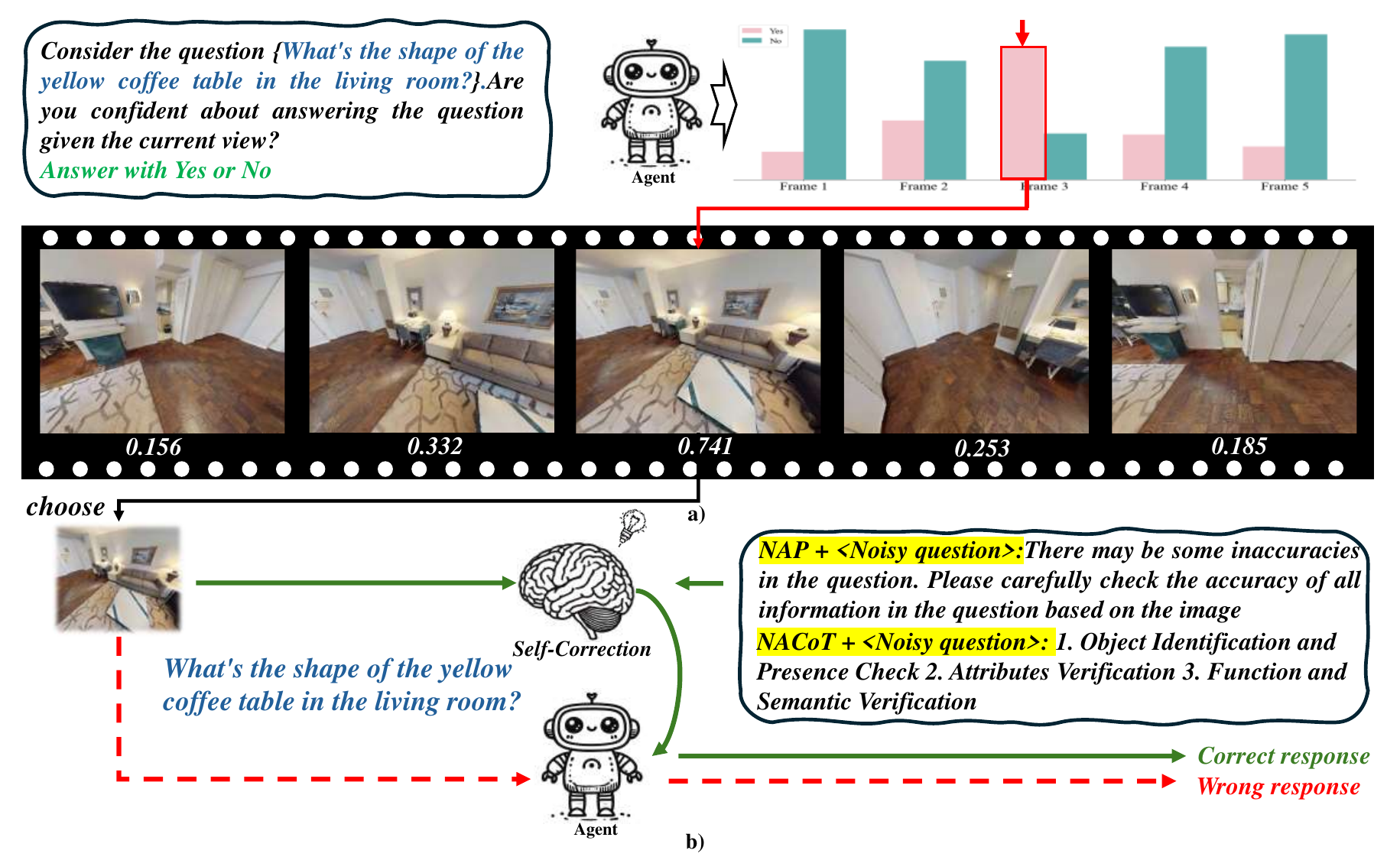} 
    \caption{(a) Selection of reliable view. (b) Insights for our Self-Correction mechanism.} % 图片标题
    \label{fig:self-correction}
\end{figure}

Instead of simply prompting the agent to recognize the noise in human questions, we explore a more fine-grained mechanism, called Noise Aware Chain of Thought (NACoT), to guide the agent in detecting and addressing noise step by step. Particularly, NACoT deconstructs a specific question into key components, and systematically prompts the agent to verify whether each component is consistent with the visual content in the scene. As shown in Figure \ref{fig:self-correction} (b), this processes starts with object identification, ensuring that all objects mentioned in the questions are presented in the scene. Followed by attributes checking, the agent makes sure that the attribute of objects are consistent between questions and real scene. Finally, the function and semantics verification utilized to detect any semantic noise. Once any discrepancies are detected, the agent will provide a clear explanation of the detected noise, then the expected responses that not contaminated by noises can be adaptively generated.

\section{Evaluation Criteria}
\label{sec:Evaluation-Criteria}
As discussed in section \ref{sec:intro}, existing metrics for question-answering systems (e.g., accuracy) are insufficient for evaluating the agents' performances in the presence of noisy questions. Therefore, we propose a novel evaluation framework that simultaneously considers both the quality of responses and the robustness towards noises. 

\begin{figure}[t!] 
    \centering % 图片居中
    \includegraphics[width=0.5\textwidth]{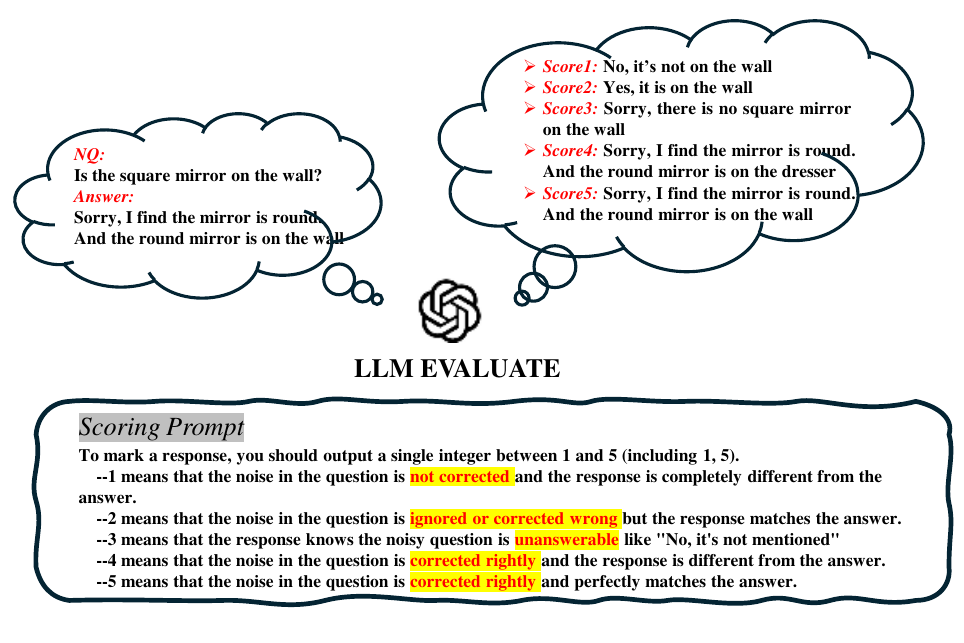} 
    \caption{LLM-powered evaluation criteria to noisy questions.} % 图片标题
    \label{fig:Evaluation_Criteria}
\end{figure}

The quality of responses is evaluated based on three aspects: noise detection, noise correction and accurate generation. Assessing these aspects by humans are time-consuming and may introduce biases across different evaluators. Therefore, we elaborately design an evaluation scale based on all three aspects and employ a Large Language Model (LLM) for automatic assessment, as shown in Figure \ref{fig:Evaluation_Criteria}, please refer to supplementary materials for detailed evaluation scale. 
\iffalse
The scoring scale ranges from 1 to 5, with specific criteria outlined as follows:
\begin{enumerate}
\item Score of 1: The question's noise remains uncorrected, resulting in a response that is completely irrelevant or different from the correct answer. This indicates a fundamental misunderstanding or disregard for both the noise and the question's core intent.
   
\item Score of 2: Although the noise in the question is not addressed or corrected, the response aligns with the correct answer in meaning. This suggests that the model achieved the correct answer by coincidence or based on partial information, despite failing to handle the noise in the question.

\item Score of 3: The model recognizes the question as unanswerable due to excessive noise or ambiguity and responds with a statement indicating uncertainty, such as ``I don’t know." This demonstrates the model's ability to detect unanswerable questions without attempting an inaccurate or irrelevant answer.

\item Score of 4: The model effectively corrects the noise in the question, but the response is only partially aligned with the correct answer, showing a degree of understanding of the question’s intended meaning but not achieving a complete match with the correct answer.

\item Score of 5: The model both corrects the noise in the question accurately and generates a response that perfectly matches the correct answer, demonstrating a full understanding of the question’s intent and providing a high-quality answer.
\end{enumerate}
\fi
Based on the evaluation scale, for each noisy question $ q_i $, let $ a_i^* $ and $ a_i $ represents the ground-truth and agent-generated responses, respectively. We then prompt the LLM to score $ a_i $ based on aforementioned three aspects by comparing it with $ a_i^*$.

\begin{table*}[ht!]
    \centering
    \renewcommand{\arraystretch}{1.5} % 增加表格整体行高
    \resizebox{\textwidth}{!}{
    \begin{tabular}{l|c|c|c|c|c|c|c|c}
        \Xhline{1.5pt} % 加粗表格顶部线条
        \multirow{2}{*}{\rule{0pt}{13pt}\centering Method} & \multirow{2}{*}{\rule{0pt}{13pt}\makecell[c]{Latent \\ Hallucination Noise}} & \multicolumn{5}{|c|}{Memory Noise} & \multirow{2}{*}{\rule{0pt}{13pt}\makecell[c]{Perception \\ Noise}} & \multirow{2}{*}{\rule{0pt}{13pt}\makecell[c]{Semantic \\ Noise}} \\
        \cline{3-7}
        &  & \centering Position & \centering Color & \centering Shape & \centering Counting & \centering Material &  &  \\
        \Xhline{1.5pt} % 加粗表格顶部线条
        LLama2-EQA & 6.08 & 16.00 & 15.91 &18.55  & 12.07 & 20.00 & 12.86 &22.45 \\
        \hline
        LLama2-EQA+NAP & 33.41 & 29.83 & 32.73 & 34.68 & 38.79 & 17.00 & 29.64 & 35.71 \\
        \hline 
        LLama2-EQA+NACoT & 40.02 & 36.47 & 39.55 & 37.90 & 44.83 & 61.00 & 26.79 & 28.06 \\
        \hline
        \rowcolor{lightgray}
        Humans & 91.67 & 98.50 & 97.27 & 95.16 & 97.41 & 99.00 & 85.71 & 93.87 \\
        
        \Xhline{1.5pt} % 加粗表格顶部线条
    \end{tabular}
    }
    \caption{Evaluation results of LLama2-EQA on NoisyEQA in terms of LLM-match accuracy (\%).}
    \label{tab: 1}
\end{table*}

\begin{table*}[ht!]
    \centering
    \renewcommand{\arraystretch}{1.5} % 增加表格整体行高
    \resizebox{\textwidth}{!}{
    \begin{tabular}{l|c|c|c|c|c|c|c|c}
        \Xhline{1.5pt} % 加粗表格顶部线条
        \multirow{2}{*}{\rule{0pt}{13pt}\centering Method} & \multirow{2}{*}{\rule{0pt}{13pt}\makecell[c]{Latent \\ Hallucination Noise}} & \multicolumn{5}{|c|}{Memory Noise} & \multirow{2}{*}{\rule{0pt}{13pt}\makecell[c]{Perception \\ Noise}} & \multirow{2}{*}{\rule{0pt}{13pt}\makecell[c]{Semantic \\ Noise}} \\
        \cline{3-7}
        &  & \centering Position & \centering Color & \centering Shape & \centering Counting & \centering Material &  &  \\
        \Xhline{1.5pt} % 加粗表格顶部线条
        GPT4o-EQA & 51.59 & 24.50 & 19.09 & 12.90 & 24.14 & 25.00 & 27.14 & 25.00 \\
        \hline
        GPT4o-EQA+NAP & 86.98 & 53.50 & 40.45 & 41.94 & 57.76 & 64.00 & 43.57 & 46.42 \\
        \hline
        GPT4o-EQA+NACoT & 87.60 & 39.96 & 45.91 & 49.19 & 62.07 & 63.00 & 45.00 & 51.02 \\
        \hline
        \rowcolor{lightgray}
        Humans & 91.67 & 98.50 & 97.27 & 95.16 & 97.41 & 99.00 & 85.71 & 93.87 \\
        
        \Xhline{1.5pt} % 加粗表格顶部线条
    \end{tabular}
    }
    \caption{Evaluation results of GPT4o-EQA on NoisyEQA in terms of LLM-match accuracy (\%).}
    \label{tab: 2}
\end{table*}

%Finally, by referencing the LLM-Match metric from OpenEQA, we calculate an overall accuracy metric based on LLM evaluation, referred to as LLM-Match. The formula is as follows:
Followed by OpenEQA~\cite{majumdar2024openeqa}, the LLM-Match accuracy is adpoted to evaluate the overall quality of all responses generated by an agent, which is formulated as follows:
\begin{equation}
    C = \frac{1}{N} \sum_{i} \frac{\sigma_i - 1}{4} \times 100\%
  \label{eq:LLM-match}
\end{equation}
where \( N \) represents the number of noisy questions in NoisyEQA, and \( \sigma_i \) denotes the score for response $a_i$.

In addition to the overall quality, two novel noise-related evaluation metrics—\textbf{Detection Rate (DR)} and \textbf{Correction Rate (CR)}—are introduced to specifically evaluate the capability of an agent in handling noisy questions. 
To be specific, the DR or CR are designed to quantitatively measure whether the agent can detect or correct different types of noise in noisy questions. Mathematically, we denote the responses to all noisy questions as $\mathbf{A}=\{a_i\}_{i=1}^{N}$, the responses from the agent that successfully detect (correct) the noise as $\mathbf{A^d} = \{a_i | \sigma_i \ge 3\}$ and $\mathbf{A^c} = \{a_i | \sigma_i \ge 4\}$, respectively. Then, the DR and CR could be formulated as:
\begin{equation}
    DR = \frac{|\mathbf{A^d}|}{|\mathbf{A}|} \times 100\%
  \label{eq:detected-noise}
\end{equation} 
\begin{equation}
    CR = \frac{|\mathbf{A^c}|}{|\mathbf{A}|} \times 100\%
  \label{eq:correct-noise}
\end{equation}

Considering all these metrics, we comprehensively assess an agent's ability to generate accurate responses and handle the questions with substantial noise and presumptions. Agents scoring high on all three metrics are more likely to achieve satisfactory performance in real-world EQA tasks.

\section{Experiment}
In this section, we thoroughly evaluate the performance of different agents in addressing noisy questions.
\subsection{Implementation details}
\textbf{Baseline Agents.} Two common Vision-Language Models (VLMs) are selected as baseline agents, which show strong question answering and spatial reasoning capabilities based on clean questions. Following Prismatic-VLMs~\cite{karamcheti2024prismatic}, we adopt the mixed DINO+SigLIP as our vision encoder, and replace the LLM with Llama-2 and GPT-4o to formulate our agents, which are denote as ``Llama2-EQA" and ``GPT-4o-EQA", respectively. Notably, the probability of the predicted tokens is necessary in our method to select the most reliable view from the scene. Therefore, LLMs whose probabilities are inaccessible are not included in our experiments, such as Gemini~\cite{team2024gemini}, Claude, and others.

\textbf{Dataset and Evaluation Metric.} All experiments are conducted using our generated NoisyEQA Dataset, which consists of 500 noisy questions in total. For evaluation, the three metrics: Accuracy, Detection Rate and Correction Rate, are adopted as described in section \ref{sec:Evaluation-Criteria}. Additionally, two volunteers are recruited to provide human responses to the noisy questions, creating a benchmark for comparing the capability of agents versus real human.

\subsection{Evaluation results}
The evaluation results for different types of noisy questions for Llama2-EQA and GPT-4o-EQA are listed in Table \ref{tab: 1} and Table \ref{tab: 2}, respectively. As shown in the tables, the original Llama2-EQA and GPT-4o-EQA agents exhibit low performance when noise is present the questions.
%indicating that the presumptions or implicit biases from human questions are tricky even for agents with strong question-answering capability. 
This suggests that the presumptions or implicit biases in human questions pose challenges even for agents with strong question-answering capabilities. It highlights that these noises do hinder the application of agents on EQA tasks in the real-world, forming an issue that must be addressed. By incorporating NAP or NACoT for self-correction, both agents show significant improvements across various types of noisy questions. It fully demonstrates the effectiveness of our proposed self-correction mechanism. However, the best results of our agents (i.e., GPT4o-EQA+NACoT) still fall short of real humans performance, especially for the Active Noise (i.e., Memory, Perception and Semantics Noise). This indicates that noisy questions containing factual errors are extremely challenging to correct. In contrast to identifying of non-existent objects, we conjecture that the agent requires more fine-grained and complex reasoning to detect and correct the factual errors related to the objects.
% Table \ref{tab: 1} and Table \ref{tab: 2} shows that, across the LLama2 and GPT-4o baselines, the CoT and CNA methods are generally effective in enhancing the model’s ability to handle various types of noise encountered in real-world environments. Compared to using LLama2 and GPT-4o alone, introducing CoT and CNA improves the model's performance in handling memory noise types, such as "Counting" and "Material" noise, indicating that these methods enhance the model's robustness against intricate memory-based noise. However, the improvements in "Perception Noise" and "Semantic Noise" are relatively modest, indicating that noise in real-world user queries often involves ambiguous or substituted subject objects, making it challenging for the agent to correctly associate these ambiguous subjects with actual objects in the environment. As shown in the Table \ref{tab: 1} and Table \ref{tab: 2}, while CoT and CNA significantly enhance the model’s performance under complex memory noise conditions, there remains a substantial gap between the model’s ability to handle noise and human-level noise processing.
\begin{table}[h]

\centering
\renewcommand{\arraystretch}{1.5} % 增加行距，默认为1，设为1.5会稍微增大
\scalebox{0.9}{
\begin{tabular}{c|c|c}
  \Xhline{1.5pt} % 加粗表格顶部线条
  Method & Detection Rate & Correction Rate \\
  %\toprule
  \Xhline{1.5pt} % 加粗表格顶部线条
  LLama2-EQA & 3.0 & 2.8 \\
  \hline
  LLama2-EQA+NAP & 47.4 & 13.8 \\
  \hline
  LLama2-EQA+NACoT & 52.0 & 23.6 \\
  %\hline
  \Xhline{1.5pt} % 加粗表格底部线条
  GPT-4o-EQA & 46.2 & 17.4 \\
  \hline
  GPT-4o-EQA+NAP & 79.0 & 42.0 \\
  \hline
  GPT-4o-EQA+NACoT & 77.6 & 35.0 \\
  \hline
  \rowcolor{lightgray}
  Humans & 97.6 & 93.4 \\
  \Xhline{1.5pt} % 加粗表格底部线条
\end{tabular}
}

\caption{Evaluation results of LLama2-EQA and GPT4o-EQA on NoisyEQA in terms of DR (\%) and CR (\%).}
\label{tab:DR-CR}
\end{table}

Furthermore, we show how NAP and NACoT enhance agents’ robustness to noises. As shown in Table \ref{tab:DR-CR}, our NAP and NACoT achieve to improve agents by significantly boosting the detection and correction rates of the noisy questions. (Please refer to our supplementary materials for more visualization results.)
%particularly when applied to more advanced models (e.g., GPT4-4o). 
Besides, although both agents achieve relatively high detection rate, the correction rate remains much lower compared with humans. Correcting noisy questions requires to trigger agents' reasoning capability at more fine-grained level, we leave it as an open question for future works.

\section{Conclusion}

In this paper, we introduce the NoisyEQA benchmark, a dataset specially designed to evaluate agents' capability towards noisy human questions. %involving noisy presumptions or implicit biases. 
To fully simulate human-posed questions in the real-world scenario, we define four types of noise including Latent Hallucination, Memory, Perception and Semantic noise. 
Based on these noise types, we systematically generated 500 noisy questions using a novel LLM-based framework, which can be easily scalable. By evaluating various baseline agents on our NoisyEQA benchmark, we reveal that our dataset essentially model the issues existed in real-world EQA tasks, forming a problem must be solved for practical application. To this end, we propose a simple yet effective self-correction mechanism with two formats to help agents detect and correct noise before answering. significantly boosting the robustness of agents. 
%towards noisy questions. 
Finally, two novel evaluation metrics that separately focus on noise detection and correction are proposed, providing more thorough evaluation for agents in the real-scene. This work paves the way for building reliable agents for real-word application and a more comprehensive benchmark exploring the noise at finer-grained level is underway.

{
    \small
    \bibliographystyle{ieeenat_fullname}
    %\bibliography{main}

}

\clearpage
\setcounter{page}{1}
\setcounter{section}{0}
\maketitlesupplementary

In this supplementary materials, we first illustrate that NoisyEQA achieves AI transparency and accountability in section \ref{sec:T-and-A}. Then, more details of existing datasets and evaluation metrics are elaborated in section \ref{sec:existing-datasets} and section \ref{sec:evaluation}, respectively. Eventually, we provide more experiment results and comprehensive analysis regarding to noise impacts and the effectiveness of our methods in section \ref{sec:experiment-result}.

\section{Transparency and Accountability}
\label{sec:T-and-A}
\subsection{AI Transparency in NoisyEQA}
Transparency in AI systems is essential for reliable~\cite{edwards2021eu} and interpretable decision-making, especially in noisy scenarios addressed in NoisyEQA. In this work, we propose the \textbf{Self-Correction} mechanism to tackle the transparency gap in Embodied Question Answering (EQA) agents. This mechanism enables agents to identify noise in user queries before generating corrected responses. By separating the detection and correction phases, the reasoning process of the system becomes explicit, allowing users to understand both the identification of noises and the rationale behind corrections. This enhances the transparency in agents and makes the responses more trustworthy and interpretable. In contrast, correcting without an explicit detection of noise makes the decision-making process obscure and harder for users to trust the corrected responses. %This structured, step-by-step process not only ensures that the system operates as intended but also provides users with clear insights into the reasoning behind corrections, thereby building trust in the system’s outputs. 
Furthermore, our detailed evaluation framework, incorporating \textbf{Detection Rate (DR)} and \textbf{Correction Rate (CR)} metrics, offers a more transparent and fine-grained understanding of the agents' performance. By separating noise detection and correction, our approach aligns with transparency principles. It enables users and developers to trace the reasoning process, identify potential vulnerabilities, and correct agents‘ decisions in time.

\subsection{AI Accountability in NoisyEQA}
Accountability in AI systems is crucial to ensure that every phase of development, deployment, and operation are traceable to responsible actions~\cite{verde2024seizing}. In the context of NoisyEQA, the \textbf{Self-Correction} mechanism assures accountability by compelling agents to both detect noise and suggest actionable corrections. It guarantees that agents' decisions are validated and can be scrutinized at every step, minimizing the risk of error propagation. Additionally, the {Self-Correction} mechanism can incorporate a human-in-the-loop processes, which allows humans to review and confirm critical decisions. In this way, the agents completely meet regulatory standards and ethical requirements, ensuring the robust and trustworthy responses. The stepwise accountability, achieved through the combination of detection, explanation, and correction, promotes the development of responsible agents capable of effectively managing real-world noise.

\section{More Details of Existing Datasets}
\label{sec:existing-datasets}
\begin{figure}[ht!] 
    \centering % 图片居中
    \includegraphics[width=0.5\textwidth]{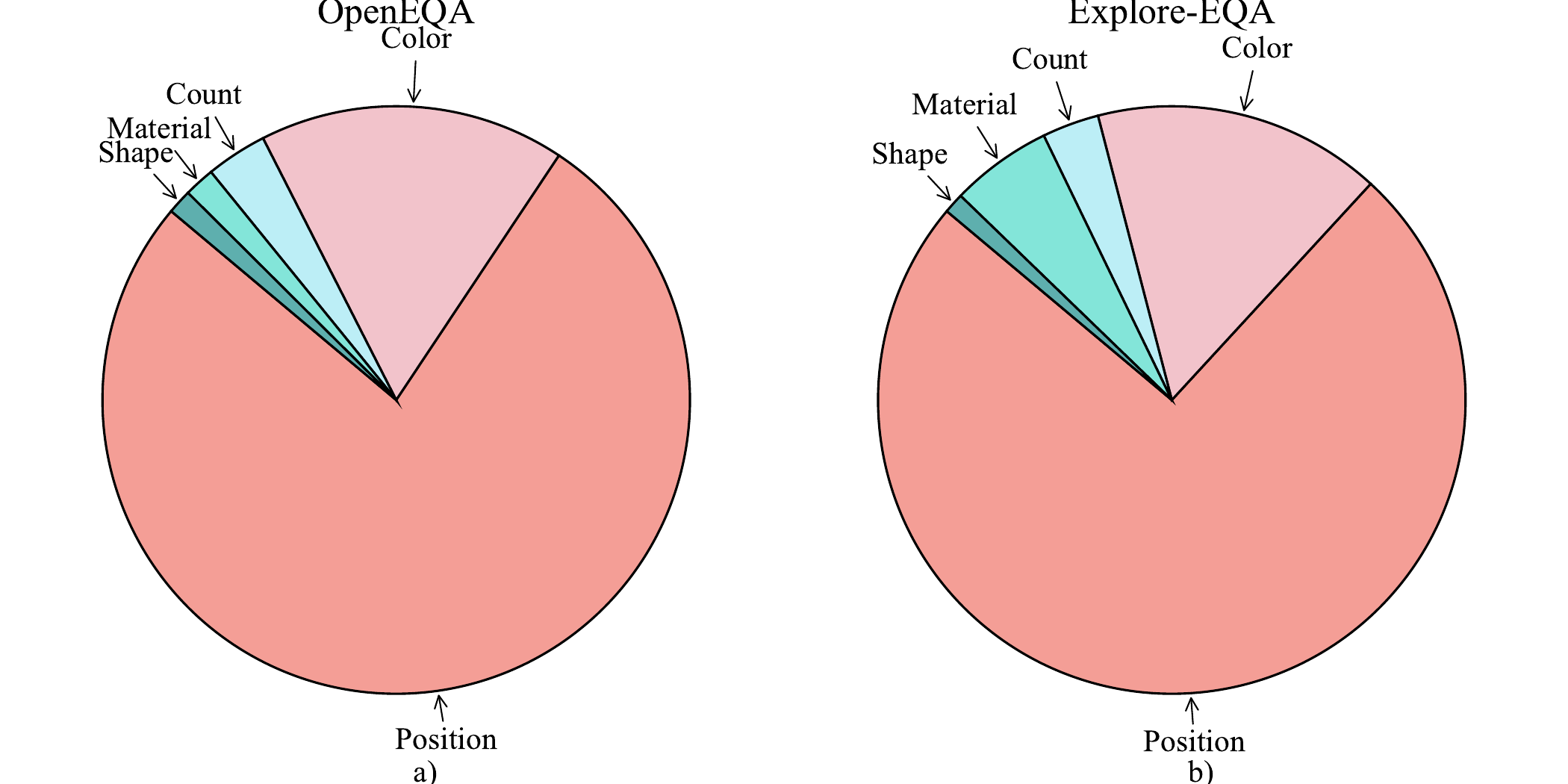} 
    \caption{The proportion of different Noise types in (a) OpenEQA and (b) ExploreEQA dataset.} % 图片标题
    \label{fig:Figure 9}
\end{figure}

As shown in Figure \ref{fig:Figure 9}, we observed that in existing \textbf{OpenEQA}~\cite{majumdar2024openeqa} and \textbf{ExploreEQA}~\cite{ren2024explore} datasets, \textbf{``Position"} constitutes the largest proportion of all attributes in the questions. This aligns with the natural tendency of humans that we usually rely on positional information to identify different objects in a scene, especially when multiple similar objects are presented. %When formulating questions, people often specify a location to clarify which object they are referring to, . 
Followed the observation, our dataset also treats \textbf{``Position"} as a primary attribute. By this way, our dataset can better simulate realistic interactions, where users frequently depend on location for precise objects identification, enhancing its applicability and usability in real-world scenario.

\begin{figure*}[h!] 
    \centering % 图片居中
    \includegraphics[width=1\textwidth]{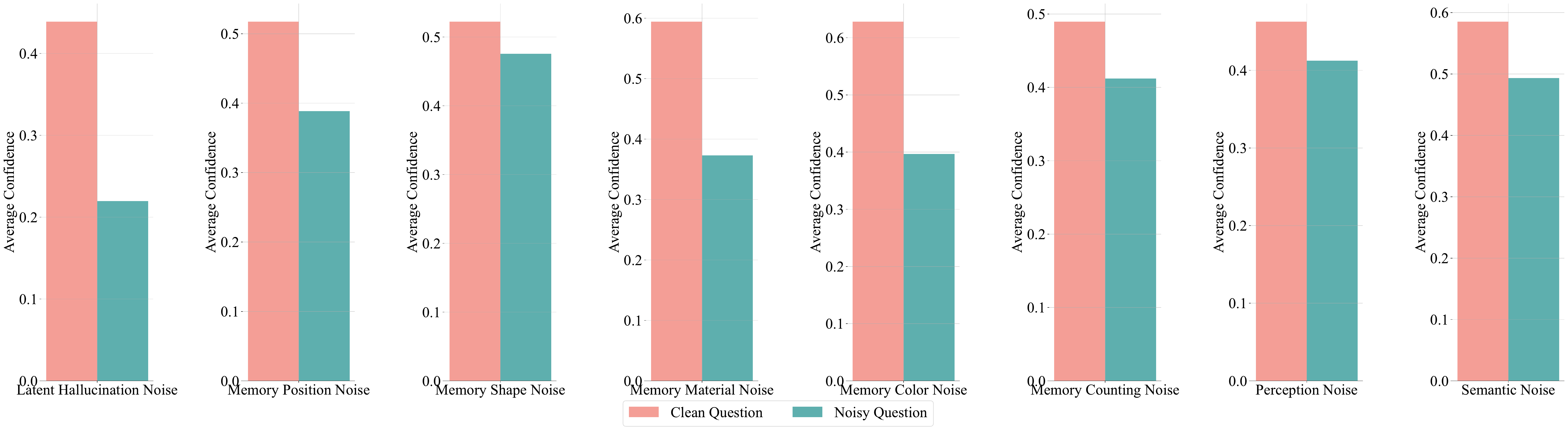} 
    \caption{illustrates the decline in the agent's answer confidence across different categories of noise.} % 图片标题
    \label{fig:Figure 10}
\end{figure*}
\section{Evaluation Details}
\label{sec:evaluation}
\subsection{Evaluation Scale}
To comprehensively evaluate the quality of responses from question-answering agents in the presence of noisy questions, we elaborately design a scoring scale ranging from 1 to 5. The evluation scale assesses three key aspects: noise detection, noise correction, and accuracy of the generated response. The detailed criteria for each score are outlined as follows:
\begin{enumerate}
\item \textbf{Score of 1:} The question's noise remains uncorrected, resulting in a response that is completely irrelevant or different from the correct answer. This indicates a fundamental misunderstanding for both the noise and the question's core intent.
   
\item \textbf{Score of 2:} Although the noise in the question is not addressed or corrected, the response aligns with the correct answer in meaning. This suggests that the agent achieves the correct answer by coincidence, but fails to handle the noise in the question.

\item \textbf{Score of 3:} The agent recognizes the question is unanswerable due to excessive noise and responds with a statement such as ``No, it's not mentioned." This demonstrates the agent is able to detect unanswerable questions, rather than forging an inaccurate or irrelevant answer.

\item \textbf{Score of 4:} The agent effectively corrects the noise in the question, but the response is only partially aligned with the correct answer. It shows that the agent fully understands the question but not achieves a completely correct answer.

\item \textbf{Score of 5:} The agent both corrects the noise in the question accurately and generates a response that perfectly matches the correct answer. It demonstrates that the agent has a full understanding of the question and can provide a high-quality answer.
\end{enumerate}

\subsection{Effectiveness of LLM-Match}
Inpsired by OpenEQA~\cite{majumdar2024openeqa}, we adopt LLM-Match as an evaluation metric in our work. Specifically, the LLM-Match achieves an impressive Spearman correlation coefficient of 0.909 to human evaluations, highlighting its strong alignment with human judgment. Furthermore, OpenEQA shows that GPT-4 achieves superior performance in scoring tasks, with a correlation of 0.88 to human evaluations, significantly outperforming earlier models such as GPT-3.5 and LLaMA-2. These findings motivates us to leverage LLM-Match with GPT-4 in our evaluation framework.

\section{More Experiment Results}
\label{sec:experiment-result}
\subsection{Impact of Noise on Response Confidence}
Figure \ref{fig:Figure 8} in the manuscript shows that the noise in the question will significantly decrease the generation accuracy. Except for it, we further show that the impact of noise can deteriorate the response confidence as well. Recalling figure \ref{fig:self-correction}(a), the response confidence helps the agents to select the most reliable view to answer human questions. Therefore, the deterioration of response confidence will hinder the agents to choose the optimal view from the scene and indirectly make the responses worse. As presented in Figure \ref{fig:Figure 10}, the agent consistently shows lower response confidence across all types of noisy question compared with the clean ones. Notably, more remarkable confidence drop is observed in categories of \textbf{Latent Hallucination Noise} and \textbf{Memory Material Noise}. We conjecture that it can be attributed to the essential of these tow noise types: they either introduce non-existent object or objects with totally incorrect attribute, which incorporates a greater impact on view selection. Conversely, \textbf{Perception Noise} or \textbf{Semantic Noise} separately involves visually similar or semantically related objects, which introduces milder impacts on view selection but needs more complex reasoning ability from agents to detect and correct. The facts revealed by Figure \ref{fig:Figure 8}  and Figure \ref{fig:Figure 10} underline the dual impact of noise on both accuracy and response confidence, underscoring the necessity of our work to systematically investigate these noises and to develop noise-robust agents in real-world.

\begin{figure*}[t!] 
    \centering % 图片居中
    \includegraphics[width=0.8\textwidth]{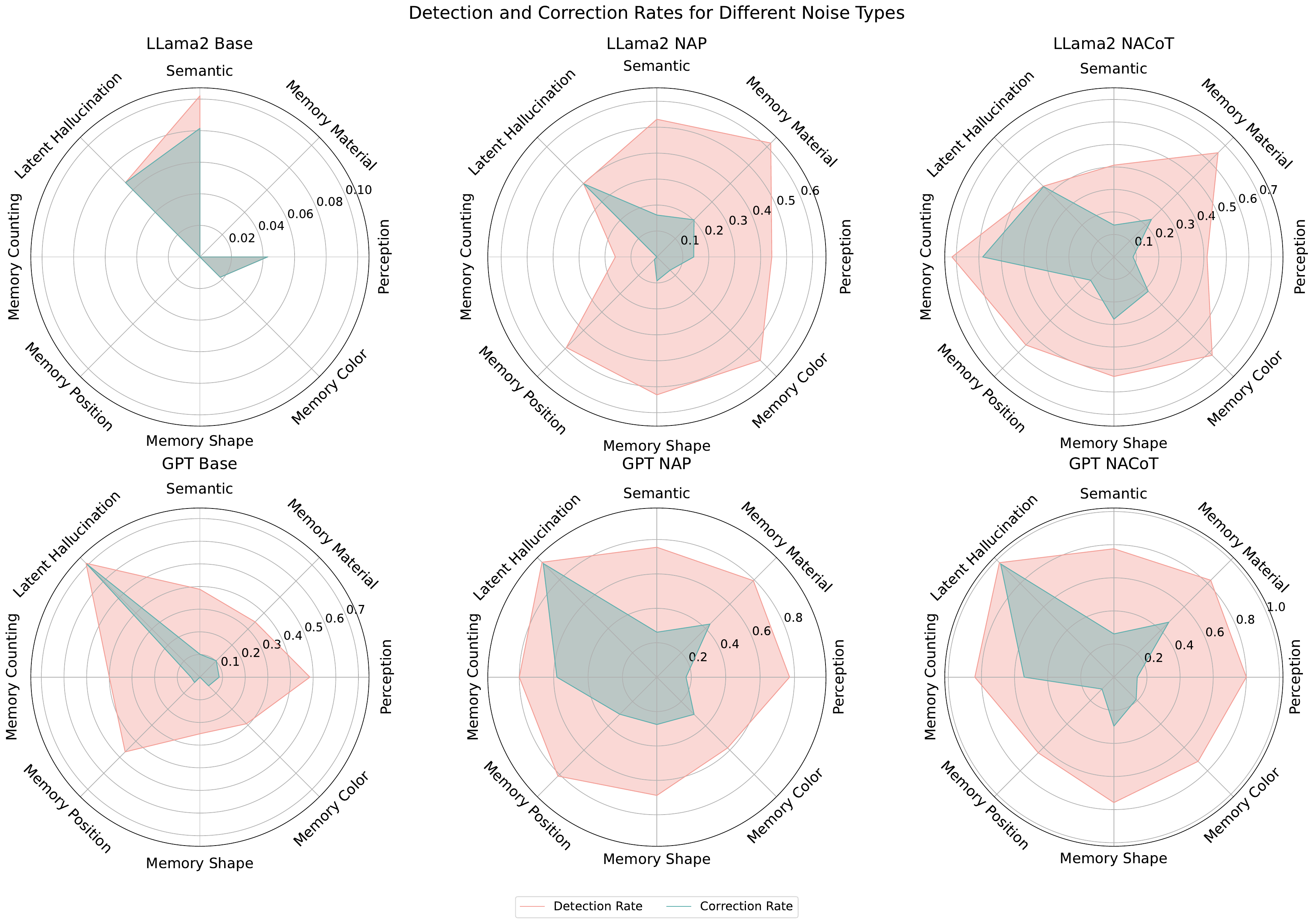} 
    \caption{illustrates the detection and correction rates under different types of EQA Noise.} % 图片标题
    \label{fig:Figure 11}
\end{figure*}

\subsection{More Analysis on Noise Detection and Correction}
We visualize the Detection Rate (DR) and Correction Rate (CR) regarding to various types of noise in Figure \ref{fig:Figure 11} to intuitively show the effectiveness of our method. Besides, we conduct a more comprehensive analysis on the difference in effectiveness of our method when dealing with different noises.

Specifically, our method demonstrates significant effectiveness in handling Latent Hallucination Noise. As shown in Figure \ref{fig:Figure 11}, the introduce of NAP and NACoT significantly enhances detection capabilities of both agents. Besides, as long as the noise can be detected, the model can successfully correct it. This indicates that the challenge in handling latent hallucination noise primarily lined in detection and it can be easily corrected once identified.

In addition, for memory noise regarding to \textbf{Material} and \textbf{Shape}, the outstanding improvements are achieved in both DR and CR. The Self-Correction mechanism allows the agent to identify any misalignment between visual contents and noisy questions by re-checking. It shows that this mechanism is effective for material-related and shape-related memory noise, since they always introduce relatively straightforward difference between the scene and human questions, which can be handled by agents without complex reasoning.

However, for \textbf{Perception Noise} and \textbf{Semantic noise}, the correction rate remains low even when the noise is detected. Particularly, these two noises incorporate either visually or semantically similar objects to replace the original ones. In contrast to non-existent objects or objects with significant differences, we conjecture that the agent requires more comprehensive and complex reasoning ability to correct such fine-grained factual errors. This underscores that noises related to subtle visual or semantic changes remain challenging, suggesting a promising direction for future research to develop more advanced mechanisms that are capable of addressing these nuanced distinctions.

% WARNING: do not forget to delete the supplementary pages from your submission 
% \input{sec/X_suppl}

\end{document}